\documentclass[letterpaper,10pt,conference]{ieeeconf}
\IEEEoverridecommandlockouts
\usepackage{amsmath,amssymb,bm}
\usepackage{graphicx,multirow,booktabs,array}
\let\labelindent\relax
\usepackage{enumitem}
\usepackage{xcolor}
\usepackage{cite}
\usepackage[hidelinks]{hyperref}
\usepackage{caption}
\usepackage{placeins}

\newif\ifanonymous
\renewcommand{\keywords}[1]{\par\smallskip\noindent{\small\textbf{Index Terms---}#1}\par\medskip}
\newcommand{\ours}{VEGA}

\title{\LARGE\bf VEGA: Learning Navigation VLAs from In-the-Wild Egocentric Video with Geometric Trajectory Supervision}
\vspace{-1ex}
\ifanonymous
\author{Anonymous authors}
\else
\author{Gershom Seneviratne, Yohan Abeysinghe, Jianyu An, Vaibhav Shende, Rahul Kumar, Dinesh Manocha \\ University of Maryland, College Park}
\fi
\IEEEaftertitletext{\vspace{-2ex}
\begin{minipage}{\textwidth}
    \centering
    \includegraphics[width=\linewidth]{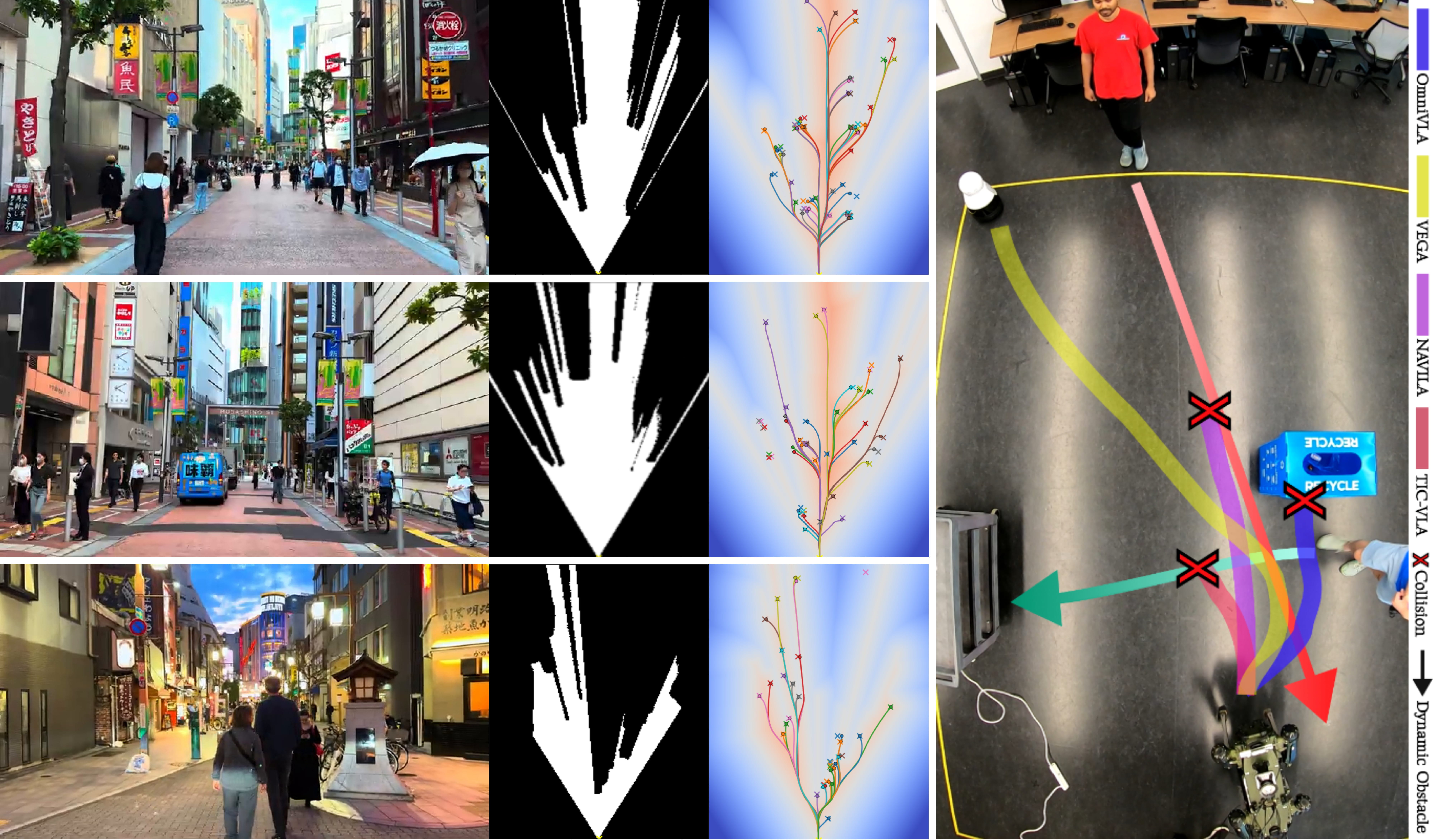}
    \vspace{-1em}
    \captionsetup{type=figure}
    \captionof{figure}{\small
    \textbf{VEGA: learning a navigation vision--language--action (VLA) model from in-the-wild egocentric video.}
    \textbf{Left:} Qualitative results of the VEGA trajectory-generation pipeline. Egocentric RGB observations are converted into visibility-aware BEV maps, with free space in white and occupied or unknown regions in black, and then into ESDFs for geometry-aware trajectory generation. Warmer ESDF regions indicate larger obstacle clearance, while cooler regions indicate unsafe space. VEGA generates multiple goal-conditioned trajectories from each observation, including paths to detected object goals and sampled free-space goals.
    \textbf{Right:} \textbf{Real-world goal-conditioned navigation with VEGA.} Scenario~4 requires navigating toward a specified goal among two dynamic and two static obstacles. Using only RGB observations and a goal at deployment, VEGA achieves 95\% success with 5\% collisions over 20 trials, whereas OmniVLA, TIC-VLA, and NaVILA achieve no successful trials, showcasing video-derived geometric supervision in cluttered, dynamic environments.
    }
    \label{fig:pipeline_results}
    \label{fig:teaser}
    \label{fig:scenario4}

\end{minipage}

\par\bigskip
\vspace{-1ex}
}
\begin{document}
\maketitle
\thispagestyle{empty}
\pagestyle{empty}
\begin{abstract}
We introduce \textbf{VEGA}, an approach for training navigation Vision-Language-Action (VLA) models from unlabeled egocentric navigation videos.
Internet-scale egocentric videos provide a scalable source of navigation-relevant visual observations, capturing cluttered scenes, close-range obstacles, and natural human motion through real-world spaces.
However, these videos are not directly usable for policy learning because they do not provide obstacle-aware trajectories conditioned on explicit navigation goals in the robot's coordinate frame.
VEGA addresses this gap by reconstructing local scene geometry from monocular video, sampling navigation goals (represented as text, image, or spatial waypoints) and generating obstacle-aware trajectories using the constructed geometry.
The resulting trajectory distribution is then used to train a flow-matching VLA navigation policy.
By using geometry exclusively during training, VEGA distills obstacle-aware planning directly into a vision-based policy. 
Furthermore, we introduce \textbf{VEGA-Bench}, a benchmark containing 250k scenes and approximately 8 million navigation goals paired with scene geometry, designed to evaluate goal progress, collision avoidance, and obstacle clearance of VLAs. 
In real-world trials, VEGA improves success by at least 65 percentage points, reduces collisions by at least 61.5\%, and improves obstacle clearance by at least 51.4\% relative to the evaluated baselines.
On VEGA-Bench, VEGA achieves competitive goal progress while reducing collisions by 18.2\% and improving obstacle clearance by 22.1\% over the strongest baseline under MoGe-2-based evaluation.
We further validate VEGA in dynamic Isaac-Sim scenarios with moving pedestrians and static obstacles, providing additional evidence of obstacle-aware navigation under dynamic conditions.
VEGA achieves these results with supervision from just 250 hours of egocentric video, approximately 2.6\% of the strongest navigation baseline's training-data duration, without collecting robot demonstrations.
This demonstrates the scalability and effectiveness of video-derived geometric supervision for training obstacle-aware navigation VLAs.
% The code and benchmark will be released at the time of publication.
\end{abstract}


\vspace{-2ex}

% Our approach is motivated by the fact that internet-scale egocentric videos provide a scalable source of navigation-relevant visual observations across diverse real-world environments including cluttered scenes and close-range obstacles. 

\section{Introduction}
\label{sec:introduction}

Vision-Language-Action (VLA) models map visual observations and language instructions directly to robot actions, offering a promising route toward general-purpose robot control by transferring semantic knowledge from vision-language pretraining into embodied action policies~\cite{brohan2023rt2visionlanguageactionmodelstransfer,intelligence2025pi_,kim2024openvla,nvidia2025gr00tn1openfoundation}.
Their recent progress has been most visible in manipulation, where large task-oriented robot datasets spanning diverse embodiments have enabled scalable policy training \cite{intelligence2025pi_, o2024open, black2024pi_0, kawaharazuka2025vision, kim2024openvla}. 
In contrast, ground mobile robot navigation lacks comparably large-scale demonstrations in complex navigation scenarios, making it difficult to scale navigation VLA training in the same way as manipulation \cite{karnan2022scand, liang2025gndglobalnavigationdataset, shah2023gnm}.
% In contrast, ground mobile robot navigation has far fewer large-scale demonstrations than manipulation, and existing datasets either remain relatively small or focus on specific settings such as social navigation, rather than providing broad goal-conditioned, obstacle-aware trajectories across cluttered everyday environments \cite{karnan2022scand, liang2025gndglobalnavigationdataset, shah2023gnm}. 
While autonomous driving datasets provide large-scale navigation data \cite{caesar2020nuscenes, chang2019argoverse, sun2020scalability}, they target structured driving scenarios and are not designed to capture the close-range obstacle avoidance, object-centric goals, and indoor--outdoor clutter encountered by general mobile robots.
This limited availability of suitable training data makes it difficult to scale navigation VLA training in the same way as manipulation, motivating methods that convert widely available egocentric videos into supervision for goal-conditioned mobile robot navigation.

Existing robot navigation datasets provide useful demonstrations for visual goal reaching, and cross-embodiment policy learning~\cite{karnan2022scand,shah2023gnm,liang2025gndglobalnavigationdataset}, but they offer limited coverage of navigation toward object goals in dense, close-range clutter, where safe behavior depends not only on the goal, but also on nearby obstacles, and available free space. Furthermore, these datasets provide sparse goal-conditioned supervision: a recorded trajectory typically demonstrates motion toward a single implicit or specified destination, while the same scene may contain multiple semantically meaningful and physically reachable goals.
% , including people, doorways, furniture and other objects, as well as intermediate viewpoints.
Therefore, training a broadly goal-conditioned navigation policy requires denser supervision than a single recorded path through a largely unobstructed environment. Such supervision must capture how feasible trajectories vary with the chosen goal and surrounding obstacles while maintaining traversability and obstacle clearance.
Without sufficient coverage of obstacle-aware trajectories toward diverse goals, a policy may rely on dominant traversal patterns in the demonstrations, limiting its ability to reach goals that require different routes through clutter~\cite{glossop2025castcounterfactuallabelsimprove, seneviratne2026chop,zhang2025creste}.

% Collecting teleoperated data to learn this behavior scales poorly with the required coverage of environments, goals, obstacle layouts, and feasible paths.
Classical navigation systems explicitly reason about free space and obstacle clearance using geometric maps, but typically require online mapping and may rely on additional depth or LiDAR sensors. These systems also lack the inherent open-vocabulary semantic reasoning capabilities of VLAs~\cite{werby2024hierarchical,inglin2026less}.
This motivates methods that combine the semantic reasoning capabilities of VLAs with the obstacle-aware geometric planning of classical methods.

To learn such behavior without costly robot data collection, we turn to egocentric videos readily available on the internet, including walking tours, bike rides, and other first-person navigation activities.
These videos capture diverse layouts, viewpoints, obstacles, traversability cues, and natural motion patterns, but lack robot action annotations, explicit navigation goals, and robot-executable trajectories.
The challenge therefore is to transform these unlabeled videos into goal-conditioned, obstacle-aware trajectory supervision for robot navigation.

We present \textbf{VEGA} (\textbf{V}ision-language navigation from \textbf{E}gocentric Video with \textbf{G}eometric \textbf{A}ction supervision), an approach for training navigation VLAs from unlabeled egocentric videos. 
VEGA reconstructs local 3D structure, and generates an obstacle-aware trajectory dataset toward scene-derived multimodal-goals; language, image, or waypoint. 
These trajectories supervise a flow-matching VLA that predicts navigation actions from RGB observations and a goal at inference time.

We also introduce \textbf{VEGA-Bench}, an evaluation suite with 250k scenes and 8 million multimodal targets for measuring goal progress, collision rate, and obstacle clearance. 

Our main contributions are summarized as follows:
\begin{itemize}
    \item \textbf{Automatic trajectory supervision from egocentric video:} We introduce VEGA, a novel method that converts unlabeled egocentric video into goal-conditioned navigation trajectory labels by reconstructing scene geometry and generating obstacle-aware paths, without manual action annotation or robot demonstrations.
    \item \textbf{Large-scale multimodal trajectory dataset:} We construct a dataset with approximately 8 million goal-conditioned trajectories aligned with language, image-region, and spatial-waypoint goal annotations, together with reconstructed 3D point clouds, Euclidean Signed Distance Fields (ESDFs), and obstacle-aware reference trajectories.

    \item \textbf{VEGA-Bench:} We introduce an evaluation tool that uses the generated scene geometry to assess trajectories proposed by navigation VLAs in terms of goal alignment, collision rate, and obstacle clearance.

    \item \textbf{Real-world validation:}
    We deploy VEGA on a physical robot, improving success by at least 65 percentage points, reducing collisions by at least 61.5\%, and increasing obstacle clearance by at least 51.4\% over state-of-the-art baselines. We further evaluate VEGA on a held-out VEGA-Bench test set and in dynamic Isaac-Sim scenarios.
\end{itemize}

These gains are achieved with navigation supervision derived from only 250 hours of egocentric video, approximately 2.6\% of the strongest navigation baseline's training-data duration.
We will release the trained model weights, dataset, and VEGA-Bench evaluation suite upon publication.

\section{Related Work}
\begin{figure*}[t]
    \centering
    \includegraphics[width=\textwidth]{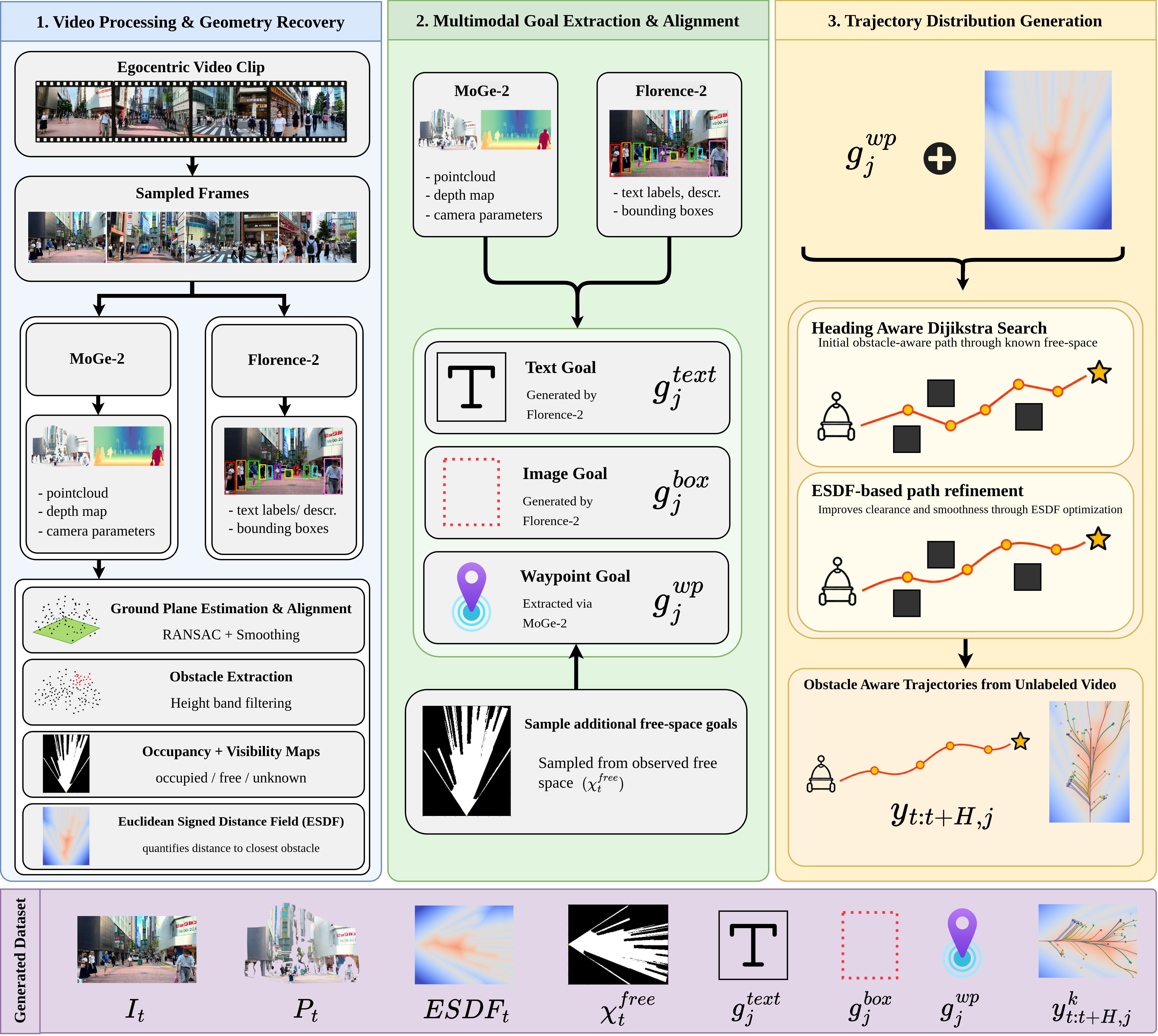}
    \caption{\footnotesize{\textbf{Overview of VEGA dataset generation.}
    Unlabeled egocentric video frames are processed using MoGe-2 to reconstruct local scene geometry and construct visibility-aware ESDFs.
    Florence-2 detections are grounded in the reconstructed point map to obtain aligned text, image-region, and waypoint goals, while auxiliary waypoint goals are sampled from observed free space.
    For each goal, heading-aware Dijkstra search generates an initial path through known free space.
    ESDF-based refinement then optimizes the interior waypoints for obstacle clearance and smoothness while keeping the endpoints fixed.
    The resulting training tuples pair RGB observations with goal annotations, modality labels, and reference waypoint trajectories.
    Reconstructed geometry is used to generate supervision and is not required at policy inference.
    }}
    \label{fig:dataset_generation_pipeline}
    \vspace{-15pt}
\end{figure*}

\subsection{Vision-Language-Action Models for Navigation}

Recent work has explored the use of VLAs for robot navigation \cite{cheng2025navilaleggedrobotvisionlanguageaction, hirose2025omnivlaomnimodalvisionlanguageactionmodel, Castro2025VAMOSAH}.
For example, NaVILA \cite{cheng2025navilaleggedrobotvisionlanguageaction} uses a hierarchical policy that converts natural-language instructions into intermediate navigation actions executed by a low-level controller. Furthermore, OmniVLA \cite{hirose2025omnivlaomnimodalvisionlanguageactionmodel} supports navigation with goals specified through natural language, goal images, or 2D poses.
% VAMOS~\cite{Castro2025VAMOSAH} decouples semantic planning from embodiment grounding using a high-level planner and an embodiment-specific affordance model.
These methods demonstrate the promise of VLA-based navigation, but rely on robot navigation datasets for action supervision.
Such datasets are expensive to collect and provide limited coverage of close-range obstacle avoidance and counterfactual goals within the same scene.
In contrast, VEGA constructs goal-conditioned, obstacle-aware navigation supervision directly from unlabeled egocentric video by reconstructing scene geometry and generating trajectories toward multiple scene-derived goals.
This enables navigation VLA training without collecting robot navigation demonstrations, allowing supervision to scale with readily available egocentric video rather than robot data collection.
% Moreover, a single observation-goal pair may admit multiple feasible trajectories, while many navigation policies are trained with a single demonstrated target trajectory.
% VEGA uses a flow-matching action head to represent continuous trajectory distributions, while its primary contribution lies in generating large-scale geometry-aware supervision from unlabeled egocentric video.

\subsection{Learning Navigation from Egocentric Video and Geometry}

Recent work has explored methods to use egocentric video as scalable supervision for navigation. 
For example, some methods use egocentric videos to learn high-level trajectories ~\cite{kumar2020learning}, while others use visual odometry on web-scale walking or driving videos to extract trajectory distributions and train visuomotor navigation policies~\cite{liu2025citywalker, Chen2025SocialNavTH}. 
Unlike these approaches, VEGA uses reconstructed scene geometry to generate collision-free trajectories to multiple goal objects within the same scene, enabling explicit goal-conditioned language supervision rather than learning only from a single observed trajectory that is not explicitly goal-conditioned. 
A related approach, LeLaN, uses a pretrained navigation policy to generate trajectories toward extracted goals in unlabeled egocentric videos, then trains a language-conditioned navigation model from these labels~\cite{hirose2024lelan}.
Because these labels are generated by a pretrained navigation policy learned from multi-robot navigation data~\cite{sridhar2024nomad}, they inherit that policy's obstacle-avoidance limitations~\cite{gode2025flownav}.
This can compromise the quality of the generated training trajectories in cluttered scenes, where safe navigation requires fine-grained reasoning about local geometry and clearance.
In contrast, VEGA generates goal-directed trajectories from reconstructed scene geometry, explicitly incorporating collision avoidance and obstacle clearance into its training supervision.

% Geometry-based representations have long supported collision checking, obstacle-clearance estimation, and motion planning, with recent work exploring signed-distance fields and neural methods for learning such fields~\cite{ortiz2022isdf,finean2021predicted,bukhari25icra, inglin2026less}. 
% Unlike these methods, which typically use geometry directly during planning or optimization at deployment, VEGA uses reconstructed scene geometry only at training time to generate obstacle-aware supervision from action-free egocentric video. 
% This allows VEGA to distill geometric planning behavior into a navigation VLA that requires only RGB observations and goal conditioning at inference, avoiding the additional sensor and compute costs associated with online depth or LiDAR-based geometric mapping.

Geometry-based representations have long supported collision checking and motion planning, with recent work exploring signed-distance fields and neural representations for planning~\cite{ortiz2022isdf,finean2021predicted,bukhari25icra}.
LiMo~\cite{inglin2026less} similarly generates goal-conditioned trajectories using robot-collected observations with co-registered elevation maps.
In contrast, VEGA reconstructs local geometry directly from monocular egocentric images to generate multimodal text-, image-, and waypoint-conditioned navigation supervision, without requiring pre-existing maps or registered robot data.
This geometric supervision is distilled into a navigation VLA that uses only RGB observations and goal conditioning at inference, avoiding online reconstruction and additional depth or LiDAR sensing.
\section{Background}
\label{sec:background}

\subsection{Vision-Language-Action Models for Navigation}

Vision-Language-Action (VLA) models adapt pretrained vision-language models to embodied control, allowing policies to map visual observations and goal instructions to robot actions while leveraging semantic knowledge from large-scale pretraining. Given an RGB observation \(I_t\) and a goal specification \(g\), a navigation VLA predicts an action chunk:
\begin{equation}
    \pi_\theta(\mathbf{a}_{t:t+H} \mid I_t, g),
    \label{eq:vla_policy}
\end{equation}
where \(\mathbf{a}_{t:t+H}\) denotes the robot's intended motion over the horizon \(H\). For mobile robot navigation, the action chunk can be represented as a path or trajectory in the robot's local frame~\cite{hirose2025omnivlaomnimodalvisionlanguageactionmodel,cheng2025navilaleggedrobotvisionlanguageaction}.

\subsection{Flow Matching for Trajectory Prediction}
\label{subsec:flow_matching}

Flow matching is a generative modeling approach that learns a vector field to transform samples from a simple prior distribution into samples from a target data distribution~\cite{lipman2022flow,liu2022flow}. 
In robot control, the target distribution can represent action chunks or future waypoint trajectories, allowing a policy to model multimodal motions rather than predicting a single deterministic output.

Let \(\mathbf{y}_1\) denote a target trajectory and \(\mathbf{y}_0 \sim \mathcal{N}(0,\mathbf{I})\) denote a noise sample of the same dimension. 
For a linear path
\begin{equation}
    \mathbf{y}_\tau = (1-\tau)\mathbf{y}_0 + \tau \mathbf{y}_1,
    \quad \tau \in [0,1],
    \label{eq:flow_interpolation}
\end{equation}
the model is trained to predict the corresponding vector field:
\begin{equation}
    \mathcal{L}_{\mathrm{FM}}(\theta)
    =
    \mathbb{E}
    \left[
    \left\|
    v_\theta(\mathbf{y}_\tau, \tau, I_t, g)
    -
    (\mathbf{y}_1 - \mathbf{y}_0)
    \right\|_2^2
    \right].
    \label{eq:flow_matching_loss}
\end{equation}
At inference time, a trajectory is generated by sampling noise and integrating the learned vector field from \(\tau=0\) to \(\tau=1\). 
This formulation allows the policy to represent multimodal distributions over trajectories conditioned on the same observation and goal.

\subsection{Monocular Geometry Reconstruction}
\label{subsec:geometric_reconstruction}

Recent monocular geometry foundation models can estimate metric 3D structure from monocular RGB frames, making it possible to recover approximate local scene geometry from videos~\cite{ren2026anydepthdepthestimationeasy, yang2024depthanythingunleashingpower, wang2025mogeunlockingaccuratemonocular, wang2025moge2accuratemonoculargeometry}. 
Given an RGB frame \(I_t \in \mathbb{R}^{H \times W \times 3}\), these methods estimate a dense point map \(P_t\), where each pixel \((u,v)\) is mapped to a 3D point in the camera coordinate frame:
\begin{equation}
P_t(u,v) =
\begin{bmatrix}
X & Y & Z
\end{bmatrix}^{T}
\in \mathbb{R}^{3}
\end{equation}

\section{Methodology}
\label{sec:methodology}

% \begin{figure}[t]
%     \centering
%     \includegraphics[width=\columnwidth]{fig/datagen.jpg}
%     \caption{Qualitative results of the VEGA trajectory-generation
%     pipeline on egocentric video frames. Each row shows a sampled
%     timestamp. The left column shows the RGB observation, the middle
%     column shows the visibility-aware BEV map with free space shown
%     in white and occupied or unknown regions shown in black, and the
%     right column overlays generated goal-conditioned trajectories
%     on the ESDF. In the ESDF visualization, warmer colors indicate
%     larger positive clearance from obstacles, while cooler colors
%     indicate unsafe regions. VEGA generates multiple trajectories
%     from the same frame, including paths to detected object goals
%     and sampled free-space goals.}
%     \label{fig:pipeline_results}
%     \vspace{-10pt}
% \end{figure}

\begin{figure}[t]
    \centering
    \includegraphics[width=\columnwidth]{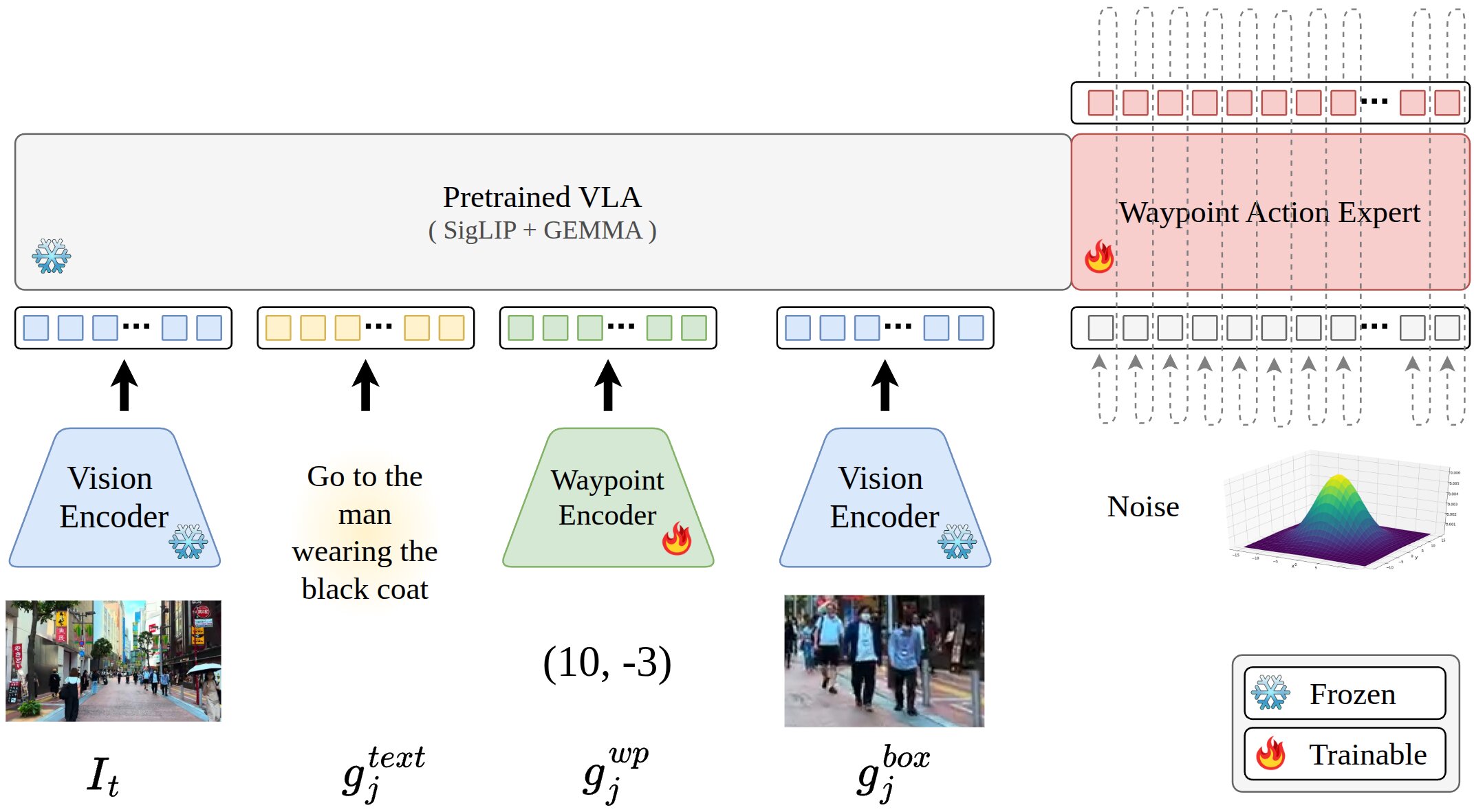}
    \caption{VEGA multimodal goal-conditioned VLA architecture.
    The current egocentric image \(I_t\), text goal
    \(g^{\mathrm{text}}_j\), waypoint goal \(g^{\mathrm{wp}}_j\),
    and image-region goal \(g^{\mathrm{box}}_j\) are encoded as
    conditioning tokens for a pretrained VLM backbone. The visual
    encoders are kept frozen, while the waypoint encoder and
    waypoint action expert are trainable. During flow-matching
    inference, noisy waypoint tokens are processed by the action
    expert, which attends to the conditioning tokens and outputs
    a continuous local waypoint trajectory.}
    \label{fig:vega_arch}
    \vspace{-4ex}
\end{figure}

\subsection{Trajectory Distribution Generation from Unlabeled Egocentric Videos}
\label{sec:trajectory_generation}

To create a target trajectory distribution that is both goal-oriented and obstacle-aware, we curate unlabeled egocentric videos from online walking tours, bike rides, and other first-person navigation footage.
Each video is uniformly sampled at a fixed frame rate to obtain RGB observations ($I_t \in \mathbb{R}^{H \times W \times 3}$), where each sampled frame is treated as a local navigation scene.
Fig.~\ref{fig:dataset_generation_pipeline} summarizes the data annotation pipeline.

\subsubsection{Geometry Recovery and Visibility-Aware ESDF Construction}
\label{methodology:esdf}

For each frame \(I_t \in \mathbb{R}^{H \times W \times 3}\), VEGA reconstructs a dense metric point map \(P_t(u,v) \in \mathbb{R}^3\) in the camera frame using MoGe-2 \cite{wang2025moge2accuratemonoculargeometry}. We fit a local ground plane to the reconstructed points using RANSAC~\cite{RANSAC10.1145/358669.358692}, and rotate the point map so that this plane is aligned with the robot-frame (xy)-plane, with the (z)-axis pointing upward. We define a local bird's-eye-view domain ($\Omega_{\mathrm{BEV}}$) around the robot and project the leveled point map onto this ground plane. Obstacle points are extracted by height filtering:
\begin{equation}
    \mathcal{P}_{\mathrm{obs}}
    =
    \left\{
    \mathbf{p}
    \mid
    z_{\min} \leq \mathbf{p}_z \leq z_{\max},
    \ \mathbf{p}_{x,y} \in \Omega_{\mathrm{BEV}}
    \right\}.
\end{equation}
and rasterize $\mathcal{P}_{\mathrm{obs}}$ to obtain the BEV occupancy grid $\mathcal{M}_{\mathrm{occ}}$. We also raytrace through the organized point map to construct a visibility mask ($\mathcal{M}_{\mathrm{vis}}$), allowing the BEV map to distinguish known free space ($\mathcal{X}_{\mathrm{free}}$), occupied space ($\mathcal{X}_{\mathrm{occ}}$), and unobserved space ($\mathcal{X}_{\mathrm{unk}}$).

Finally, VEGA computes a Euclidean Signed Distance Field $\mathrm{ESDF}(x,y)$ over the local frame:
\begin{equation}
\mathrm{ESDF}(x,y)=
\begin{cases}
+d_{\mathrm{unsafe}}(x,y), & (x,y) \in \mathcal{X}_{\mathrm{free}}, \\
-d_{\mathrm{free}}(x,y), & (x,y) \in \mathcal{X}_{\mathrm{occ}} \cup \mathcal{X}_{\mathrm{unk}},
\end{cases}
\end{equation}
where $d_{\mathrm{unsafe}}(x,y)$ is the Euclidean distance to the nearest occupied or unobserved cell in $\mathcal{X}_{\mathrm{occ}}\cup\mathcal{X}_{\mathrm{unk}}$, and $d_{\mathrm{free}}(x,y)$ is the distance to the nearest known-free cell.
Positive ESDF values therefore measure clearance within known free space, while negative values denote occupied or unobserved regions that the planner should avoid.

\subsubsection{Multimodal Goal Anchoring}
VEGA uses Florence-2~\cite{xiao2023florence2advancingunifiedrepresentation} to extract object labels and bounding boxes from each RGB frame $I_t$. For each detected object $j$, we query the corresponding region in the MoGe-2 point map and project it onto the robot-centric ground plane to obtain a waypoint goal:
\begin{equation}
g^{\mathrm{wp}}_j = (x_j, y_j) \in \mathbb{R}^2 .
\end{equation}
This yields aligned text, image-region, and waypoint goals $(g^{\mathrm{text}}_j, g^{\mathrm{box}}_j, g^{\mathrm{wp}}_j)$ for each detected target. VEGA also samples auxiliary waypoint goals from the observed free-space subset $\mathcal{X}_{\mathrm{free}}$ to provide supervision for arbitrary traversable locations.

% \subsubsection{MPPI-Based Trajectory Generation}
% \label{methodology:mppi}

% For each waypoint goal $g^{\mathrm{wp}}_j \in \mathbb{R}^2$, VEGA uses a Model Predictive Path Integral (MPPI) planner to generate a target waypoint trajectory $\mathbf{y}^{*}_{t:t+H,j}$. Candidate trajectories are rolled out under non-holonomic robot dynamics and scored using ESDF-based goal-reaching, collision, clearance, control-effort, and smoothness costs. The lowest-cost collision-free rollout is used as the reference trajectory; if the goal is not reachable within the local map, VEGA selects the best collision-free partial trajectory that makes progress toward the goal. We elaborate on the MPPI implementation in  Appendix \ref{app:mppi}.

% \gershom{remove mentions of MPPI}

\subsubsection{ESDF-Based Trajectory Generation}
\label{methodology:esdf_planner}

For each waypoint goal $g^{\mathrm{wp}}_j \in \mathbb{R}^2$, VEGA first uses a heading-aware Dijkstra search over the visibility-aware ESDF to generate a collision-free, goal-directed path from the robot origin. If an in-map goal is not directly reachable, VEGA selects the nearest reachable endpoint; goals outside the local ESDF are discarded. The resulting path is refined on the continuous ESDF with fixed start and end points by optimizing its interior waypoints:
\begin{equation}
\begin{aligned}
\min_{\{p_i\}_{i=1}^{H-2}} \quad
& \lambda_{\mathrm{obs}}\sum_{i=1}^{H-2}
\left[\max\!\left(0, d_{\mathrm{safe}}-\mathrm{ESDF}(p_i)\right)\right]^2 \\
& {}+ \lambda_{\mathrm{smooth}}\sum_{i=1}^{H-2}
\|p_i-\bar{p}_i\|_2^2.
\end{aligned}
\end{equation}
where $\bar{p}_i$ is a weighted average of neighboring waypoints. This objective moves points away from obstacles until sufficient clearance is reached while preserving local path smoothness. The resulting smooth, collision-aware trajectory is used as the reference waypoint trajectory $\mathbf{y}^{*}_{t:t+H,j}$.

\subsubsection{Generated Training Tuples}

After trajectory generation, each sampled frame $I_t$ yields multiple goal-conditioned training tuples:
\begin{equation}
\mathcal{D}_t =
\{
(
I_t,
g_j,
m_j,
\hat{\mathbf{y}}_{t:t+H,j}
)
\}_{j=1}^{K_t},
\label{eq:vega_training_tuple}
\end{equation}
where $K_t$ is the number of goals associated with frame $I_t$, $g_j$ is the $j$-th goal, $m_j \in \{\mathrm{text}, \mathrm{box}, \mathrm{wp}\}$ denotes the goal modality, and $\hat{\mathbf{y}}_{t:t+H,j}$ is the
corresponding reference waypoint trajectory.

\subsection{Multimodal Goal-Conditioned VLA}

As shown in Fig.~\ref{fig:vega_arch}, VEGA adapts the $\pi_{0.5}$ VLA architecture~\cite{black2024pi_0,intelligence2025pi_}, using the base checkpoint pretrained SigLIP vision encoder, Gemma language backbone, and flow-matching action expert to predict robot-centric 2D waypoint chunks $\hat{\mathbf{y}}_{t:t+H} \in \mathbb{R}^{H \times 2}$. The waypoint encoder and action expert are initialized from scratch.

Inspired by OmniVLA~\cite{hirose2025omnivlaomnimodalvisionlanguageactionmodel}, VEGA supports text, image-region, and waypoint goals. The current RGB observation and image-region goals are encoded using SigLIP, while text goals are processed by the language backbone. A lightweight trainable MLP maps 2D waypoint goals into the VLM's token space. The observation and goal tokens form a shared prefix context that conditions the waypoint action expert.

We train the waypoint encoder and action expert while keeping the SigLIP encoder and Gemma backbone frozen. Training uses the flow-matching objective in Eq.~\ref{eq:flow_matching_loss}, with ESDF-generated reference trajectories as target samples and Gaussian noise as source samples. At inference, the action expert integrates the learned vector field to generate waypoint trajectories conditioned on the current RGB observation and specified goal, without requiring scene reconstruction or an ESDF.

% \subsection{Flow-Matching Training Objective}

% Let \(\mathbf{y}_1 = \mathbf{y}^{*}_{t:t+H}\) denote the target trajectory generated by MPPI, and let \(\mathbf{y}_0 \sim \mathcal{N}(0,I)\) denote a noise trajectory. For a sampled flow time \(\tau \in [0,1]\), we define
% \[
%     \mathbf{y}_\tau = (1-\tau)\mathbf{y}_0 + \tau \mathbf{y}_1 .
% \]
% The action head is trained to predict the vector field from the noisy trajectory toward the target trajectory:
% \[
%     \mathcal{L}_{\mathrm{FM}}
%     =
%     \mathbb{E}_{\mathbf{y}_0,\mathbf{y}_1,\tau}
%     \left[
%     \left\|
%     v_\theta(\mathbf{y}_\tau,\tau,I_t,g)
%     -
%     (\mathbf{y}_1-\mathbf{y}_0)
%     \right\|_2^2
%     \right].
% \]
% During training, the goal conditioning \(g\) is sampled from the aligned goal modalities when available. This trains the same waypoint generation head to operate under language, image, and waypoint goal conditioning.

% \subsection{Inference}

% At inference time, VEGA receives an RGB observation and a goal specification. Starting from a noise trajectory, the model integrates the learned flow field to generate a waypoint trajectory:
% \[
%     \hat{\mathbf{y}}_{t:t+H}
%     =
%     \pi_\theta(I_t,g).
% \]
% The predicted waypoints are tracked by a low-level controller. Depth estimation, point cloud generation, ESDF construction, Florence-2 annotation, and MPPI planning are used only for dataset generation and are not required during deployment.

\section{Analysis and Results}

Our evaluation addresses three questions:
\textbf{(1)} Can video-derived supervision enable reliable goal reaching and obstacle avoidance on a physical robot?
\textbf{(2)} How does VEGA compare with existing VLAs on trajectory quality and multimodal goal reaching?
\textbf{(3)} How reliable is the generated geometric supervision, and how robust are VEGA's performance gains to the geometry source and collision criteria?

\begin{figure*}[t]
    \centering

    % ==================== Left: Real-world qualitative results ====================
    \begin{minipage}[t]{0.57\textwidth}
        \vspace{0pt}
        \centering

        \includegraphics[width=\linewidth]{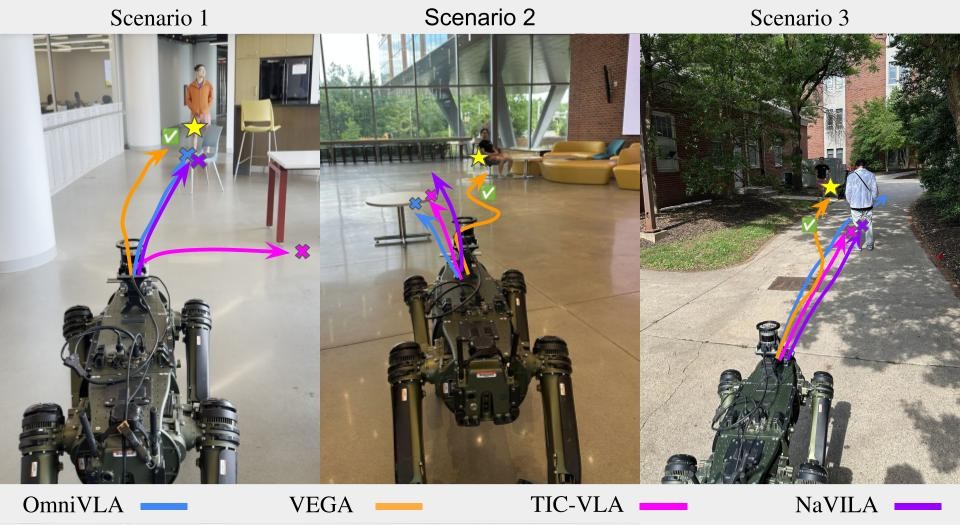}

        \captionof{figure}{
        \textbf{Real-world goal-conditioned navigation with VEGA.}
        Qualitative comparison across three real-world scenarios with static and dynamic obstacles.
        Each trajectory is directed toward the yellow-star goal; check marks indicate success and X marks indicate collisions or failures.
        VEGA reaches the goal while avoiding obstacles, whereas baseline VLAs often collide, stop short, or miss the target.
        }
        \label{fig:comps}
    \end{minipage}
    \hfill
    % ==================== Middle: Isaac-Sim scenarios ====================
    \begin{minipage}[t]{0.19\textwidth}
        \vspace{0pt}
        \centering

        \includegraphics[width=\linewidth]{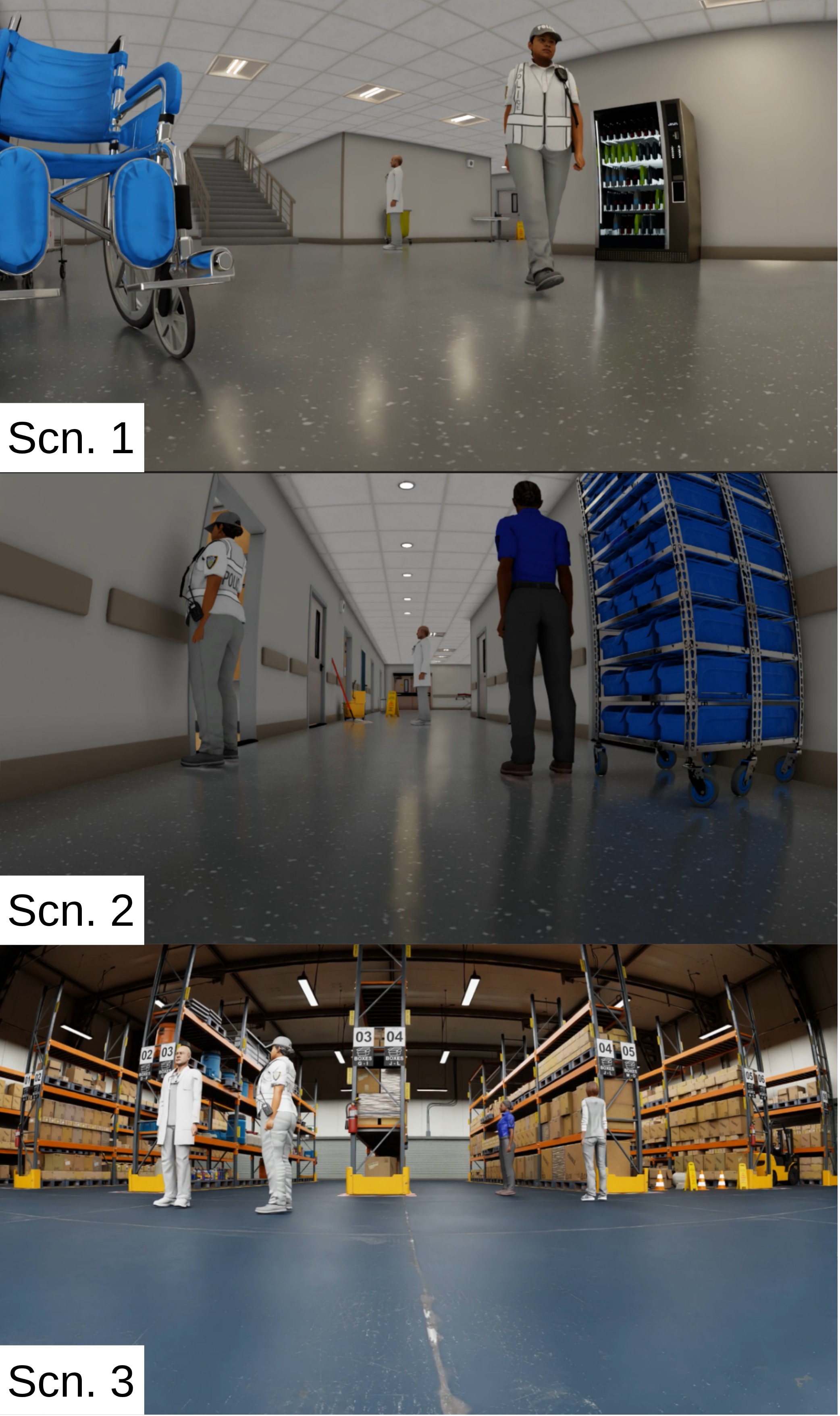}

        \captionof{figure}{
        \textbf{Isaac-Sim scenarios.}
        Robot-view snapshots for Scenarios~1--3; the stationary person in white is the goal.
        }
        \label{fig:isaac_scenarios}
    \end{minipage}
    \hfill
    % ==================== Right: Isaac-Sim quantitative results ====================
    \begin{minipage}[t]{0.22\textwidth}
        \vspace{0pt}
        \centering

        \scriptsize
        \setlength{\tabcolsep}{1.5pt}
        \renewcommand{\arraystretch}{1.22}

        \begin{tabular}{llrrr}
        \toprule
        Scene & Method & S$\uparrow$ & C$\downarrow$ & Cl$\uparrow$ \\
        \midrule
        Scn1 & VEGA & \textbf{1.00} & \textbf{0.00} & \textbf{0.85} \\
           & OmniVLA     & 0.42 & 0.58 & 0.04 \\
           & NaVILA      & 0.80 & \textbf{0.00} & 0.84 \\
           & TIC-VLA     & 0.18 & 0.50 & 0.45 \\
        \midrule
        Scn2 & VEGA & \textbf{0.90} & \textbf{0.10} & 0.49 \\
           & OmniVLA     & 0.06 & 0.96 & 0.00 \\
           & NaVILA      & \textbf{0.90} & 0.02 & \textbf{0.53} \\
           & TIC-VLA     & \textbf{0.90} & 0.04 & 0.48 \\
        \midrule
        Scn3 & VEGA & \textbf{1.00} & \textbf{0.00} & 1.06 \\
           & OmniVLA     & 0.04 & 0.16 & 0.13 \\
           & NaVILA      & 0.00 & 0.00 & 1.52 \\
           & TIC-VLA     & 0.74 & 0.06 & 0.61 \\
        \bottomrule
        \end{tabular}

        \captionof{table}{
        Isaac-Sim results over 50 trials per method and scenario.
        $S$ and $C$ are rates; $Cl$ is clearance in meters. \ours{} outperforms baselines overall in 50 simulations per scenario.
        }
        \label{tab:isaac_sim}
    \end{minipage}

    \vspace{-4ex}
\end{figure*}
\subsection{Experimental Setup}

\noindent \textbf{Baselines.}
We compare VEGA with OmniVLA~\cite{hirose2025omnivlaomnimodalvisionlanguageactionmodel}, TIC-VLA~\cite{huang2026tic}, and NaVILA~\cite{cheng2025navilaleggedrobotvisionlanguageaction}, three state-of-the-art navigation VLAs.
This comparison assesses whether geometric supervision derived from unlabeled Internet videos enables effective goal-directed navigation and obstacle avoidance relative to VLAs trained using robot demonstrations.
Goal specifications are provided through each model's supported interface while preserving the intended target across methods.
For real-world execution, policy outputs are passed to a shared control stack~\cite{seneviratne2026chop} that provides no additional path planning or obstacle avoidance, but allows all methods to replan from updated observations before completing a prediction.
NaVILA's high-level navigation commands are converted into equivalent local waypoints and executed through this shared stack.
We do not use NaVILA's native LiDAR-conditioned obstacle-avoiding locomotion policy, since our goal is to evaluate obstacle avoidance encoded in the VLA predictions themselves rather than corrections introduced by downstream navigation modules.
Thus, all methods are evaluated under a common execution interface without additional obstacle avoidance.

\paragraph{Implementation Details.}
During dataset construction, we use a $25\,\mathrm{m}\times20\,\mathrm{m}$ BEV at $0.02\,\mathrm{m}$/cell resolution, centered on the robot's forward workspace. We use a safety distance of $d_{\mathrm{safe}}=0.5\,\mathrm{m}$, corresponding to the robot radius, and generate waypoint chunks with horizon $H=50$. Goal reaching is defined as terminating within 0.5\,m of the target.

\noindent \textbf{Metrics.}
We evaluate each method along three axes:
\begin{itemize}[noitemsep,topsep=1pt]
    \item \textbf{Goal reaching.}
    On VEGA-Bench, we report average normalized goal progress, defined as the fractional reduction in goal distance normalized by the initial robot-goal distance.
    In real-world trials, success requires the robot to reach within a threshold distance of the intended goal without any collision or safety intervention.

    \item \textbf{Collision avoidance.}
    On VEGA-Bench, collision rate is the fraction of trajectories that enter occupied, unknown, or out-of-bounds ESDF regions.
    In real-world trials, collision rate is the fraction of trials with physical contact or safety intervention.

    \item \textbf{Obstacle clearance.}
    We report the average minimum distance to the nearest obstacle along non-colliding trajectories.
\end{itemize}

\begin{table}[t]
\vspace{2ex}
\centering
% \footnotesize
\setlength{\tabcolsep}{3pt}
\renewcommand{\arraystretch}{1.1}

\resizebox{\columnwidth}{!}{%
\begin{tabular}{@{}l*{12}{c}@{}}
\toprule
& \multicolumn{3}{c}{Scenario 1}
& \multicolumn{3}{c}{Scenario 2}
& \multicolumn{3}{c}{Scenario 3}
& \multicolumn{3}{c}{Scenario 4} \\
\cmidrule(lr){2-4}
\cmidrule(lr){5-7}
\cmidrule(lr){8-10}
\cmidrule(lr){11-13}
Method
& S$\uparrow$ & C$\downarrow$ & Cl$\uparrow$
& S$\uparrow$ & C$\downarrow$ & Cl$\uparrow$
& S$\uparrow$ & C$\downarrow$ & Cl$\uparrow$
& S$\uparrow$ & C$\downarrow$ & Cl$\uparrow$ \\
\midrule
OmniVLA
& 0.15 & 0.85 & 0.35
& 0.0 & 0.65 & 0.32
& 0.0 & 0.5 & 0.44
& 0 & 1.0 & N/A \\
TIC-VLA
& 0.0 & 1.0 & N/A
& 0.1 & 0.85 & 0.37
& 0.0 & 1.0 & N/A
& 0 & 1.0 & N/A \\
NaVILA
& 0.0 & 1.0 & N/A
& 0.0 & 0.8 & 0.35
& 0.0 & 0.45 & 0.31
& 0 & 0.75 & 0.25 \\
\textbf{VEGA}
& \textbf{0.95} & \textbf{0.05} & \textbf{0.93}
& \textbf{0.75} & \textbf{0.25} & \textbf{0.56}
& \textbf{1.0} & \textbf{0.0} & \textbf{0.94}
& \textbf{0.95} & \textbf{0.05} & \textbf{0.52} \\
\bottomrule
\end{tabular}%
}

\caption{Real-world navigation over 20 trials.
S: success rate; Cl: clearance in meters, with N/A when
all trials collide; C: average collision rate
for Scenarios 1--4.}
\label{tab:real_results}
\label{tab:scenario4}
\vspace{-2ex}
\end{table}

\begin{table}[t]
\centering
\footnotesize
\setlength{\tabcolsep}{3pt}
\renewcommand{\arraystretch}{1.1}

\resizebox{\columnwidth}{!}{%
\begin{tabular}{@{}l*{9}{c}@{}}
\toprule
& \multicolumn{3}{c}{VEGA-Bench}
& \multicolumn{3}{c}{Collision composition}
& \multicolumn{3}{c}{Goal conditioning} \\
\cmidrule(lr){2-4}
\cmidrule(lr){5-7}
\cmidrule(lr){8-10}
Method
& G$\uparrow$ & C$\downarrow$ & Cl$\uparrow$
& $C_{\mathrm{occ}}$ & $C_{\mathrm{unk}}$ & $C_{\mathrm{oob}}$
& Text$\uparrow$ & Image$\uparrow$ & WP$\uparrow$ \\
\midrule
OmniVLA
& 0.88 & 6.75 & 2.35
& 71.4\% & 23.8\% & 4.74\%
& 0.8 & 0.2 & 0.6 \\
TIC-VLA
& 0.84 & 9.01 & 1.34
& 68.1\% & 26.6\% & 5.27\%
& 0.8 & N/A & N/A \\
NaVILA
& 0.77 & 7.40 & 2.18
& 68.4\% & 26.8\% & 4.9\%
& 0.6 & N/A & N/A \\
\textbf{VEGA}
& \textbf{0.89} & \textbf{5.52} & \textbf{2.87}
& 75.9\% & 20.93\% & 3.16\%
& \textbf{1.0} & \textbf{1.0} & \textbf{0.8} \\
\bottomrule
\end{tabular}%
}

\caption{VEGA-Bench evaluation, collision composition,
and goal-reaching performance by modality.
G: goal progress; C: collision percentage;
Cl: clearance in meters.
Collision composition reports percentages of collisions
in occupied ($C_{\mathrm{occ}}$), unknown ($C_{\mathrm{unk}}$),
and out-of-bounds ($C_{\mathrm{oob}}$) cells. Goal-modality results report success over 20 real-world trials per supported modality using the Scenario~4 layout with dynamic obstacles removed.;
WP denotes waypoint goals.}
\label{tab:vega_bench_results}
\label{tab:collision_decomposition}
\label{tab:goal_modality_results}
\vspace{-5ex}
\end{table}

\subsection{Real-World Robot Evaluation}

% For real-world trials, all methods are executed through the shared ROS-based stack from CHOP~\cite{seneviratne2026chop}.
% NaVILA's native action outputs are mapped to equivalent local waypoints for execution, while the other methods retain their native navigation outputs.
% The shared stack performs no additional path planning or obstacle avoidance and allows all methods to replan from updated observations before completing a prediction.

To assess real-world transfer, we deploy the methods on a quadruped in cluttered environments with both static and dynamic obstacles, as shown in Fig.~\ref{fig:comps}.
Across four scenarios with 20 trials per scenario, VEGA achieves the highest success rate and the fewest collisions among the evaluated baselines (Table~\ref{tab:real_results}). 
These results demonstrate that video-derived trajectory supervision enables effective goal reaching and obstacle avoidance on a physical robot in the tested environments.

To separately evaluate goal conditioning from dynamic-obstacle avoidance, we conduct 20 trials per supported goal modality using the Scenario~4 layout with the dynamic obstacles removed (Table~\ref{tab:vega_bench_results}).
These trials test whether each policy can interpret the specified goal and generate goal-directed motion under a simpler navigation setting.
Across text, image, and waypoint goals, VEGA produces distinct trajectories that reach the corresponding targets, while some baselines often generate similar paths despite changes in the goal specification.

\subsection{Isaac-Sim Evaluation}

We additionally evaluate navigation in three Isaac-Sim scenarios containing static obstacles and moving pedestrians.
%illustrated in Fig.~\ref{fig:isaac_scenarios}. 
Trials vary pedestrian motion onset using identical timing conditions across methods.
Following the real-world evaluation, we report success rate, collision rate, and average minimum clearance over non-collision trials in Table~\ref{tab:isaac_sim}.

\begin{table*}[!t]
\centering
\footnotesize
\setlength{\tabcolsep}{3pt}
\renewcommand{\arraystretch}{1.1}
\resizebox{\textwidth}{!}{%
\begin{tabular}{@{}l*{12}{c}@{}}
\toprule
& \multicolumn{3}{c}{OmniVLA}
& \multicolumn{3}{c}{TIC-VLA}
& \multicolumn{3}{c}{NaVILA}
& \multicolumn{3}{c}{VEGA} \\
\cmidrule(lr){2-4}
\cmidrule(lr){5-7}
\cmidrule(lr){8-10}
\cmidrule(lr){11-13}
Pointcloud source
& $G$$\uparrow$ & $C$$\downarrow$ & $Cl$$\uparrow$
& $G$$\uparrow$ & $C$$\downarrow$ & $Cl$$\uparrow$
& $G$$\uparrow$ & $C$$\downarrow$ & $Cl$$\uparrow$
& $G$$\uparrow$ & $C$$\downarrow$ & $Cl$$\uparrow$ \\
\midrule
MoGe-2
& 0.88/1.00 & 6.75 & 2.35/1.00
& 0.84/1.00 & 9.01 & 1.34/1.00
& 0.77/1.00 & 7.40 & 2.18/1.00
& 0.89/1.00 & 5.52 & 2.87/1.00 \\
UniDepthV2
& 0.87/0.97 & 6.91 & 2.31/0.95
& 0.84/0.95 & 9.43 & 1.91/0.94
& 0.76/0.96 & 7.56 & 2.14/0.94
& 0.86/0.97 & 5.89 & 2.61/0.91 \\
UniK3D
& 0.86/0.95 & 6.58 & 2.39/0.93
& 0.80/0.94 & 8.96 & 1.97/0.92
& 0.75/0.94 & 7.23 & 2.22/0.92
& 0.85/0.95 & 5.39 & 2.82/0.95 \\
Metric3Dv2
& 0.84/0.92 & 7.09 & 2.27/0.90
& 0.76/0.90 & 9.65 & 1.82/0.89
& 0.73/0.92 & 7.79 & 2.10/0.90
& 0.83/0.92 & 5.88 & 2.64/0.92 \\
% \midrule
% SCAND (LiDAR)
% & 0.37 & 5.42 & 2.64
% & 0.31 & 7.68 & 2.19
% & 0.29 & 6.13 & 2.47
% & 0.34 & 3.61 & 3.06 \\
\bottomrule
\end{tabular}%
}
\caption{Benchmark sensitivity to pointcloud source.
For goal progress ($G$) and clearance ($Cl$), each monocular-model entry reports mean $\mu$/Spearman $\rho$ relative to MoGe-2.
Collision ($C$) reports mean collision percentage.
$G$: goal progress ($\uparrow$);
$C$: collision percentage ($\downarrow$);
$Cl$: clearance in meters ($\uparrow$).}
\label{tab:geometry_sensitivity}
\vspace{-4ex}
\end{table*}

\subsection{VEGA-Bench Evaluation}

To evaluate goal-conditioned navigation at scale, we benchmark all methods on VEGA-Bench.
For the main comparison, all methods are evaluated on the common text-conditioned subset, while modality-specific real-world evaluations are reported separately.
The VEGA-Bench test set contains approximately 12.5 hours of held-out videos, with the split performed at the city-level so that frames and their derived goals and trajectories from a given video do not appear in both training and evaluation.
Predicted trajectories are evaluated against the reconstructed scene geometry using goal progress, collision rate, and obstacle clearance.
As shown in Table~\ref{tab:vega_bench_results}, VEGA achieves competitive goal progress while reducing collision rate and increasing obstacle clearance relative to the evaluated baselines.
% Additional results and clarifications from the supplied CoRL rebuttal.
\subsection{Ablations and Analysis}

\subsubsection{Sensitivity to Reconstructed Geometry}
\label{sec:geometry_validation}

Using the same reconstruction model for both supervision generation and VEGA-Bench evaluation could favor trajectories that align with its systematic biases.
To assess this, we reevaluate all policies using ESDFs and object-goal coordinates from three alternative monocular 3D models: UniDepthV2~\cite{piccinelli2025unidepthv2}, UniK3D~\cite{piccinelli2025unik3d}, and Metric3Dv2~\cite{hu2024metric3d}.
Predicted trajectories and target object identities remain fixed, while scene geometry and each target's metric location are re-estimated using the corresponding 3D model.
Goal progress is therefore recomputed using the 3D model-specific goal coordinates, while collision rate and clearance are evaluated using the corresponding ESDF.
Table~\ref{tab:geometry_sensitivity} reports mean metric values and Spearman correlations relative to evaluation with MoGe-2.
Although absolute values vary, evaluations remain strongly correlated with those obtained using MoGe-2, and method rankings are preserved.
In particular, VEGA maintains the lowest collision rate and greatest obstacle clearance across the tested reconstruction models.
These results provide evidence that VEGA's lower collision rate and greater obstacle clearance on VEGA-Bench, while having slightly better goal-reaching, are not merely artifacts of using MoGe-2 for both supervision generation and evaluation.

\subsubsection{Generated-Supervision Quality}

We first evaluate the sensitivity of the generated trajectory supervision to the monocular reconstruction model. As shown in Table~\ref{tab:dataset_sensitivity}(a), trajectories generated using UniDepthV2, UniK3D, and Metric3Dv2 retain similar goal progress and obstacle clearance, with low collision rates across all reconstruction models. This indicates that the geometric supervision pipeline is not strongly dependent on MoGe-2.
To further assess whether monocular reconstruction errors degrade the generated trajectory supervision, we compare trajectories planned from MoGe-2 reconstructions with trajectories planned from independently measured SCAND LiDAR pointclouds~\cite{karnan2022scand}.
For each matched scene and goal, both trajectories are generated using the same planner, with the geometry change,  and evaluated against the same LiDAR-ground-truth ESDF.
As shown in Table~\ref{tab:lidar_validation} (c), trajectories planned from monocular geometry achieve similar goal progress, collision rate, and obstacle clearance to those planned directly from LiDAR geometry.
The corresponding paths have a mean Fr\'echet distance of 0.28\,m, further indicating close agreement between the generated supervision.

\subsubsection{Ablation on Geometric Supervision}
To isolate the contribution of obstacle-aware planning to supervision quality, we compare policies trained with ESDF-based reference trajectories with the same policy trained with straight-line paths to the same goals.
Straight-line paths preserve the goal direction but ignore intervening obstacles and scene geometry.
As shown in Table~\ref{tab:straight_line_ablation} (b), omitting geometric planning increases collisions from 1.2\% to 14.8\% and reduces clearance from 3.12\,m to 2.14\,m, while goal progress remains comparable (0.92 vs.\ 0.94).
Because these trajectories directly supervise the policy's action predictions, unsafe or low-clearance reference trajectories provide correspondingly poorer navigation targets.
These results therefore support geometry-aware trajectory generation for improving the quality of VEGA's training supervision.

\begin{table}[t]
\vspace{2ex}
\centering
\footnotesize
\setlength{\tabcolsep}{3pt}
\renewcommand{\arraystretch}{1.1}

\begin{tabular*}{\columnwidth}
{@{\extracolsep{\fill}}lccc@{}}
\toprule

\multicolumn{4}{@{}l}{\textbf{(a) Pointcloud source for generated supervision}} \\
Source & $G\uparrow$ & $C\downarrow$ & $Cl\uparrow$ \\
\midrule
MoGe-2
& 0.97/1.00 & 0.00 & 3.35/1.00 \\
UniDepthV2
& 0.95/0.98 & 0.10 & 3.22/0.96 \\
UniK3D
& 0.94/0.97 & 0.12 & 3.27/0.98 \\
Metric3Dv2
& 0.92/0.95 & 0.18 & 3.08/0.92 \\

\specialrule{0.9pt}{4pt}{4pt}

\multicolumn{4}{@{}l}{\textbf{(b) Ablation on Geometric Supervision}} \\
Supervision & $G\uparrow$ & $C\downarrow$ & $Cl\uparrow$ \\
\midrule
ESDF-optimized & 0.92 & 1.2 & 3.12 \\
Straight-line   & 0.94 & 14.8 & 2.14 \\

\specialrule{0.9pt}{4pt}{4pt}

\multicolumn{4}{@{}l}{\textbf{(c) LiDAR-based validation of generated supervision}} \\
Planning geometry & $G\uparrow$ & $C\downarrow$ & $Cl\uparrow$ \\
\midrule
SCAND LiDAR & 0.91 & 0 & 0.492 \\
MoGe-2 & 0.881 & 3.2 & 0.385\\
\midrule
\multicolumn{3}{@{}l}{Fr\'echet distance between generated paths} & 0.28\,m \\

\bottomrule
\end{tabular*}

\caption{Ablations on generated trajectory supervision.
$G$: goal progress; $C$: collision percentage;
$Cl$: clearance in meters.
(a) Mean/Spearman correlation relative to MoGe-2.
(b) Mean metrics for ESDF-optimized and straight-line trajectory policies.
(c) Trajectories planned using either SCAND LiDAR and MoGe-2 geometry based ESDFs are evaluated against the same LiDAR-derived reference ESDF; the Fr\'echet distance measures agreement between the corresponding generated paths.}
\label{tab:dataset_sensitivity}
\label{tab:straight_line_ablation}
\label{tab:lidar_validation}
\vspace{-5ex}
\end{table}

\subsubsection{Collision Decomposition}

To distinguish avoidance of observed obstacles from conservative treatment of unknown space, we decompose collisions into occupied, unknown, and out-of-bounds (OOB) pointcloud cells (Table~\ref{tab:collision_decomposition}). The reported percentages indicate each category's share of collisions, and each colliding trajectory is categorized by the first invalid region encountered.
VEGA retains the lowest collision rate when considering occupied space alone, indicating that its advantage is not solely due to avoiding unknown regions.

\noindent Together, these analyses show that geometric planning improves supervision quality, with generated trajectories closely matching LiDAR-derived references. VEGA's performance advantage persists across the tested reconstruction models and when considering occupied-space collisions alone, supporting the reliability of the supervision and robustness of the evaluation findings.
\section{Conclusion and Limitations}

We present VEGA, a method for training goal-conditioned navigation VLAs from action-free egocentric video using geometry-derived trajectory supervision. VEGA reconstructs local scene geometry, anchors multimodal goals, and generates obstacle-aware waypoint trajectories to supervise a flow-matching policy—requiring geometric data only during training. Across VEGA-Bench and real-world robot experiments, VEGA outperforms standard VLA baselines in goal progress, collision avoidance, and obstacle clearance, demonstrating that video-derived geometric supervision is a scalable path to safe, semantically-guided navigation.

VEGA has two main limitations. First, its memoryless architecture relies solely on the current RGB frame and goal, making it vulnerable when obstacles or targets leave the field of view; integrating temporal memory or recurrent state could improve robustness under partial observability. Second, its static ESDF supervision does not explicitly predict dynamic obstacle trajectories; future work could incorporate motion prediction or time-indexed distance fields to handle moving obstacles over the planning horizon.

%We presented \textbf{VEGA}, a method for training goal-conditioned navigation VLAs from action-free egocentric video using geometry-derived trajectory supervision. 
%VEGA reconstructs local scene geometry, anchors multimodal goals, and generates obstacle-aware waypoint trajectories to supervise a flow-matching navigation policy, using geometry only during training.
%This enables Internet-scale egocentric video to be converted into useful supervision for mobile robot navigation. Across \textbf{VEGA-Bench} and real-world robot experiments, VEGA improves goal progress, collision avoidance, and obstacle clearance over the tested VLA baselines. 
%These results suggest that video-derived geometric supervision provides a scalable supervision toward obstacle-aware navigation that combine semantic goal understanding with geometric safety.

%VEGA has two main limitations. 
%First, the policy is memoryless, predicting actions from only the current RGB observation and goal conditioning, which can be brittle when goals, obstacles, or free space leave the camera field of view. 
%Temporal attention, recurrent state, or latent memory could improve robustness under partial observability. 
%Second, VEGA's local ESDF supervision does not explicitly model dynamic obstacle motion. 
%While replanning handles newly observed obstacles, the planner does not predict their future motions. Future work could incorporate trajectory prediction or time-indexed distance fields to account for moving obstacles over the planning horizon.

% \input{sec/7_limitations}
\bibliographystyle{ieeetr}
\bibliography{short_refs}
\end{document}